\documentclass[conference]{IEEEtran}
\IEEEoverridecommandlockouts
\usepackage{amsmath,amssymb,amsfonts}
\usepackage{graphicx}
\usepackage{textcomp}
\usepackage{xcolor}
\usepackage{booktabs}
\usepackage{algorithm}
\usepackage{cleveref}
\usepackage{pgfplots}
\usepgfplotslibrary{groupplots}
\usepackage[noend]{algpseudocode}

\pgfplotsset{compat=newest}
\pgfplotsset{every axis legend/.append style={legend cell align=left}}

\pgfplotsset{every axis/.append style={
                    title style={font=\small},
                    tick label style={font=\footnotesize}  
                    }}
\pgfplotsset{every axis label/.style={font=\small}}                    
\pgfplotsset{
legend image code/.code={
\draw[mark repeat=2,mark phase=2]
plot coordinates {
(0cm,0cm)
(0.15cm,0cm)        
(0.3cm,0cm)         
};%
}
}

\definecolor{pastelMagenta}{HTML}{FF48CF}
\definecolor{pastelPurple}{HTML}{8770FE}
\definecolor{pastelBlue}{HTML}{1BA1EA}
\definecolor{pastelSeaGreen}{HTML}{14B57F}
\definecolor{pastelGreen}{HTML}{3EAA0D}
\definecolor{pastelOrange}{HTML}{C38D09}
\definecolor{pastelRed}{HTML}{F5615C}

\pgfplotsset{
   	colormap={pasteljet}{
		rgb=(0.99325,0.90616,0.14394)
		rgb=(0.98387,0.90487,0.13690)
		rgb=(0.97442,0.90359,0.13021)
		rgb=(0.96489,0.90232,0.12394)
		rgb=(0.95530,0.90107,0.11813)
		rgb=(0.94564,0.89982,0.11284)
		rgb=(0.93590,0.89857,0.10813)
		rgb=(0.92611,0.89733,0.10407)
		rgb=(0.91624,0.89609,0.10072)
		rgb=(0.90631,0.89485,0.09813)
		rgb=(0.89632,0.89362,0.09634)
		rgb=(0.88627,0.89237,0.09537)
		rgb=(0.87617,0.89112,0.09525)
		rgb=(0.86601,0.88987,0.09595)
		rgb=(0.85581,0.88860,0.09745)
		rgb=(0.84556,0.88732,0.09970)
		rgb=(0.83527,0.88603,0.10265)
		rgb=(0.82494,0.88472,0.10622)
		rgb=(0.81458,0.88339,0.11035)
		rgb=(0.80418,0.88205,0.11496)
		rgb=(0.79376,0.88068,0.12001)
		rgb=(0.78331,0.87928,0.12540)
		rgb=(0.77285,0.87787,0.13111)
		rgb=(0.76237,0.87642,0.13706)
		rgb=(0.75188,0.87495,0.14323)
		rgb=(0.74139,0.87345,0.14956)
		rgb=(0.73089,0.87192,0.15603)
		rgb=(0.72039,0.87035,0.16260)
		rgb=(0.70990,0.86875,0.16926)
		rgb=(0.69942,0.86712,0.17597)
		rgb=(0.68894,0.86545,0.18272)
		rgb=(0.67849,0.86374,0.18950)
		rgb=(0.66805,0.86200,0.19629)
		rgb=(0.65764,0.86022,0.20308)
		rgb=(0.64726,0.85840,0.20986)
		rgb=(0.63690,0.85654,0.21662)
		rgb=(0.62658,0.85464,0.22335)
		rgb=(0.61629,0.85271,0.23005)
		rgb=(0.60604,0.85073,0.23671)
		rgb=(0.59584,0.84872,0.24333)
		rgb=(0.58568,0.84666,0.24990)
		rgb=(0.57556,0.84457,0.25642)
		rgb=(0.56550,0.84243,0.26288)
		rgb=(0.55548,0.84025,0.26928)
		rgb=(0.54552,0.83804,0.27563)
		rgb=(0.53562,0.83579,0.28191)
		rgb=(0.52578,0.83349,0.28813)
		rgb=(0.51599,0.83116,0.29428)
		rgb=(0.50627,0.82879,0.30036)
		rgb=(0.49661,0.82638,0.30638)
		rgb=(0.48703,0.82393,0.31232)
		rgb=(0.47750,0.82144,0.31820)
		rgb=(0.46805,0.81892,0.32400)
		rgb=(0.45867,0.81636,0.32973)
		rgb=(0.44937,0.81377,0.33538)
		rgb=(0.44014,0.81114,0.34097)
		rgb=(0.43098,0.80847,0.34648)
		rgb=(0.42191,0.80577,0.35191)
		rgb=(0.41291,0.80304,0.35727)
		rgb=(0.40400,0.80027,0.36255)
		rgb=(0.39517,0.79748,0.36776)
		rgb=(0.38643,0.79464,0.37289)
		rgb=(0.37778,0.79178,0.37794)
		rgb=(0.36921,0.78889,0.38291)
		rgb=(0.36074,0.78596,0.38781)
		rgb=(0.35236,0.78301,0.39264)
		rgb=(0.34407,0.78003,0.39738)
		rgb=(0.33588,0.77702,0.40205)
		rgb=(0.32780,0.77398,0.40664)
		rgb=(0.31981,0.77091,0.41115)
		rgb=(0.31193,0.76782,0.41559)
		rgb=(0.30415,0.76470,0.41994)
		rgb=(0.29648,0.76156,0.42422)
		rgb=(0.28892,0.75839,0.42843)
		rgb=(0.28148,0.75520,0.43255)
		rgb=(0.27415,0.75199,0.43660)
		rgb=(0.26694,0.74875,0.44057)
		rgb=(0.25986,0.74549,0.44447)
		rgb=(0.25290,0.74221,0.44828)
		rgb=(0.24607,0.73891,0.45202)
		rgb=(0.23937,0.73559,0.45569)
		rgb=(0.23281,0.73225,0.45928)
		rgb=(0.22640,0.72889,0.46279)
		rgb=(0.22012,0.72551,0.46623)
		rgb=(0.21400,0.72211,0.46959)
		rgb=(0.20803,0.71870,0.47287)
		rgb=(0.20222,0.71527,0.47608)
		rgb=(0.19657,0.71183,0.47922)
		rgb=(0.19109,0.70837,0.48228)
		rgb=(0.18578,0.70489,0.48527)
		rgb=(0.18065,0.70140,0.48819)
		rgb=(0.17571,0.69790,0.49103)
		rgb=(0.17095,0.69438,0.49380)
		rgb=(0.16638,0.69086,0.49650)
		rgb=(0.16202,0.68732,0.49913)
		rgb=(0.15785,0.68376,0.50169)
		rgb=(0.15389,0.68020,0.50417)
		rgb=(0.15015,0.67663,0.50659)
		rgb=(0.14662,0.67305,0.50894)
		rgb=(0.14330,0.66946,0.51121)
		rgb=(0.14021,0.66586,0.51343)
		rgb=(0.13734,0.66225,0.51557)
		rgb=(0.13469,0.65864,0.51765)
		rgb=(0.13227,0.65501,0.51966)
		rgb=(0.13007,0.65138,0.52161)
		rgb=(0.12809,0.64775,0.52349)
		rgb=(0.12633,0.64411,0.52531)
		rgb=(0.12478,0.64046,0.52707)
		rgb=(0.12344,0.63681,0.52876)
		rgb=(0.12231,0.63315,0.53040)
		rgb=(0.12138,0.62949,0.53197)
		rgb=(0.12064,0.62583,0.53349)
		rgb=(0.12008,0.62216,0.53495)
		rgb=(0.11970,0.61849,0.53635)
		rgb=(0.11948,0.61482,0.53769)
		rgb=(0.11942,0.61114,0.53898)
		rgb=(0.11951,0.60746,0.54022)
		rgb=(0.11974,0.60379,0.54140)
		rgb=(0.12009,0.60010,0.54253)
		rgb=(0.12057,0.59642,0.54361)
		rgb=(0.12115,0.59274,0.54464)
		rgb=(0.12183,0.58905,0.54562)
		rgb=(0.12261,0.58537,0.54656)
		rgb=(0.12346,0.58169,0.54744)
		rgb=(0.12440,0.57800,0.54829)
		rgb=(0.12539,0.57432,0.54909)
		rgb=(0.12645,0.57063,0.54984)
		rgb=(0.12757,0.56695,0.55056)
		rgb=(0.12873,0.56327,0.55123)
		rgb=(0.12993,0.55958,0.55186)
		rgb=(0.13117,0.55590,0.55246)
		rgb=(0.13244,0.55222,0.55302)
		rgb=(0.13374,0.54853,0.55354)
		rgb=(0.13507,0.54485,0.55403)
		rgb=(0.13641,0.54117,0.55448)
		rgb=(0.13777,0.53749,0.55491)
		rgb=(0.13915,0.53381,0.55530)
		rgb=(0.14054,0.53013,0.55566)
		rgb=(0.14194,0.52645,0.55599)
		rgb=(0.14334,0.52277,0.55629)
		rgb=(0.14476,0.51909,0.55657)
		rgb=(0.14618,0.51541,0.55682)
		rgb=(0.14761,0.51173,0.55705)
		rgb=(0.14904,0.50805,0.55725)
		rgb=(0.15048,0.50437,0.55743)
		rgb=(0.15192,0.50069,0.55759)
		rgb=(0.15336,0.49700,0.55772)
		rgb=(0.15482,0.49331,0.55784)
		rgb=(0.15627,0.48962,0.55794)
		rgb=(0.15773,0.48593,0.55801)
		rgb=(0.15919,0.48224,0.55807)
		rgb=(0.16067,0.47854,0.55812)
		rgb=(0.16214,0.47484,0.55814)
		rgb=(0.16362,0.47113,0.55815)
		rgb=(0.16512,0.46742,0.55814)
		rgb=(0.16662,0.46371,0.55812)
		rgb=(0.16813,0.45999,0.55808)
		rgb=(0.16965,0.45626,0.55803)
		rgb=(0.17118,0.45253,0.55797)
		rgb=(0.17272,0.44879,0.55788)
		rgb=(0.17427,0.44504,0.55779)
		rgb=(0.17584,0.44129,0.55768)
		rgb=(0.17742,0.43753,0.55756)
		rgb=(0.17902,0.43376,0.55743)
		rgb=(0.18063,0.42997,0.55728)
		rgb=(0.18226,0.42618,0.55712)
		rgb=(0.18390,0.42238,0.55694)
		rgb=(0.18556,0.41857,0.55675)
		rgb=(0.18723,0.41475,0.55655)
		rgb=(0.18892,0.41091,0.55633)
		rgb=(0.19063,0.40706,0.55609)
		rgb=(0.19236,0.40320,0.55584)
		rgb=(0.19410,0.39932,0.55556)
		rgb=(0.19586,0.39543,0.55528)
		rgb=(0.19764,0.39153,0.55497)
		rgb=(0.19943,0.38761,0.55464)
		rgb=(0.20124,0.38367,0.55429)
		rgb=(0.20306,0.37972,0.55393)
		rgb=(0.20490,0.37575,0.55353)
		rgb=(0.20676,0.37176,0.55312)
		rgb=(0.20862,0.36775,0.55268)
		rgb=(0.21050,0.36373,0.55221)
		rgb=(0.21240,0.35968,0.55171)
		rgb=(0.21430,0.35562,0.55118)
		rgb=(0.21621,0.35153,0.55063)
		rgb=(0.21813,0.34743,0.55004)
		rgb=(0.22006,0.34331,0.54941)
		rgb=(0.22199,0.33916,0.54875)
		rgb=(0.22393,0.33499,0.54805)
		rgb=(0.22586,0.33081,0.54731)
		rgb=(0.22780,0.32659,0.54653)
		rgb=(0.22974,0.32236,0.54571)
		rgb=(0.23167,0.31811,0.54483)
		rgb=(0.23360,0.31383,0.54391)
		rgb=(0.23553,0.30953,0.54294)
		rgb=(0.23744,0.30520,0.54192)
		rgb=(0.23935,0.30085,0.54084)
		rgb=(0.24124,0.29648,0.53971)
		rgb=(0.24311,0.29209,0.53852)
		rgb=(0.24497,0.28768,0.53726)
		rgb=(0.24681,0.28324,0.53594)
		rgb=(0.24863,0.27877,0.53456)
		rgb=(0.25043,0.27429,0.53310)
		rgb=(0.25219,0.26978,0.53158)
		rgb=(0.25394,0.26525,0.52998)
		rgb=(0.25565,0.26070,0.52831)
		rgb=(0.25732,0.25613,0.52656)
		rgb=(0.25897,0.25154,0.52474)
		rgb=(0.26057,0.24692,0.52283)
		rgb=(0.26214,0.24229,0.52084)
		rgb=(0.26366,0.23763,0.51876)
		rgb=(0.26515,0.23296,0.51660)
		rgb=(0.26658,0.22826,0.51435)
		rgb=(0.26797,0.22355,0.51201)
		rgb=(0.26931,0.21882,0.50958)
		rgb=(0.27059,0.21407,0.50705)
		rgb=(0.27183,0.20930,0.50443)
		rgb=(0.27301,0.20452,0.50172)
		rgb=(0.27413,0.19972,0.49891)
		rgb=(0.27519,0.19490,0.49600)
		rgb=(0.27619,0.19007,0.49300)
		rgb=(0.27713,0.18523,0.48990)
		rgb=(0.27801,0.18037,0.48670)
		rgb=(0.27883,0.17549,0.48340)
		rgb=(0.27957,0.17060,0.48000)
		rgb=(0.28025,0.16569,0.47650)
		rgb=(0.28087,0.16077,0.47290)
		rgb=(0.28141,0.15583,0.46920)
		rgb=(0.28189,0.15088,0.46541)
		rgb=(0.28229,0.14591,0.46151)
		rgb=(0.28262,0.14093,0.45752)
		rgb=(0.28288,0.13592,0.45343)
		rgb=(0.28307,0.13090,0.44924)
		rgb=(0.28319,0.12585,0.44496)
		rgb=(0.28323,0.12078,0.44058)
		rgb=(0.28320,0.11568,0.43611)
		rgb=(0.28309,0.11055,0.43155)
		rgb=(0.28291,0.10539,0.42690)
		rgb=(0.28266,0.10020,0.42216)
		rgb=(0.28233,0.09495,0.41733)
		rgb=(0.28192,0.08967,0.41241)
		rgb=(0.28145,0.08432,0.40741)
		rgb=(0.28089,0.07891,0.40233)
		rgb=(0.28027,0.07342,0.39716)
		rgb=(0.27957,0.06784,0.39192)
		rgb=(0.27879,0.06214,0.38659)
		rgb=(0.27794,0.05632,0.38119)
		rgb=(0.27702,0.05034,0.37572)
		rgb=(0.27602,0.04417,0.37016)
		rgb=(0.27495,0.03775,0.36454)
		rgb=(0.27381,0.03150,0.35885)
		rgb=(0.27259,0.02556,0.35309)
		rgb=(0.27131,0.01994,0.34727)
		rgb=(0.26994,0.01463,0.34138)
		rgb=(0.26851,0.00961,0.33543)
		rgb=(0.26700,0.00487,0.32942)
	  }
}
\pgfplotscreateplotcyclelist{pastelcolors}{%
  solid, pastelPurple, mark=none\\%
  solid, pastelBlue, mark=none\\%
  solid, pastelGreen, mark=none\\%
  solid, pastelRed, mark=none\\%
  solid, pastelMagenta, mark=none\\%
  solid, pastelOrange, mark=none\\%
  solid, pastelSeaGreen, mark=none\\%
}

\usepackage[keeplastbox]{flushend}

\usepackage[backend=bibtex,style=ieee,mincitenames=1,maxcitenames=2,maxbibnames=20]{biblatex}
\newcommand{\argmax}{\operatornamewithlimits{arg\,max}}

\def\BibTeX{{\rm B\kern-.05em{\sc i\kern-.025em b}\kern-.08em
    T\kern-.1667em\lower.7ex\hbox{E}\kern-.125emX}}
\begin{document}

\title{Learning an Urban Air Mobility\\ Encounter Model from Expert Preferences\\
}

\author{\IEEEauthorblockN{Sydney M. Katz}
\IEEEauthorblockA{\textit{Aeronautics and Astronautics} \\
\textit{Stanford University}\\
Stanford, CA 94305 \\
smkatz@stanford.edu}
\and
\IEEEauthorblockN{Anne-Claire Le Bihan}
\IEEEauthorblockA{\textit{Wayfinder Group} \\
\textit{A\textsuperscript{3} by Airbus}\\
Sunnyvale, CA 94086 \\
anneclaire.lebihan@airbus-sv.com}
\and
\IEEEauthorblockN{Mykel J. Kochenderfer}
\IEEEauthorblockA{\textit{Aeronautics and Astronautics} \\
\textit{Stanford University}\\
Stanford, CA 94305 \\
mykel@stanford.edu}
}

\maketitle

\begin{abstract}
Airspace models have played an important role in the development and evaluation of aircraft collision avoidance systems for both manned and unmanned aircraft. As Urban Air Mobility (UAM) systems are being developed, we need new encounter models that are representative of their operational environment. Developing such models is challenging due to the lack of data on UAM behavior in the airspace. While previous encounter models for other aircraft types rely on large datasets to produce realistic trajectories, this paper presents an approach to encounter modeling that instead relies on expert knowledge. In particular, recent advances in preference-based learning are extended to tune an encounter model from expert preferences. The model takes the form of a stochastic policy for a Markov decision process (MDP) in which the reward function is learned from pairwise queries of a domain expert. We evaluate the performance of two querying methods that seek to maximize the information obtained from each query. Ultimately, we demonstrate a method for generating realistic encounter trajectories with only a few minutes of an expert's time.
\end{abstract}

\section{Introduction}

As Urban Air Mobility (UAM) systems are developed, it is important to assess their safety and performance in the airspace using realistic simulations. Other safety-critical systems including the Traffic Alert and Collision Avoidance System (TCAS) \cite{TCASMonteCarlo} and its recent successor, the Airborne Collision Avoidance System X (ACAS X) \cite{ACASXMonteCarlo}, have been assessed using airspace simulations. Driving these simulations are airspace encounter models, which are probabilistic representations of typical aircraft behavior during a close encounter with another aircraft.

Encounter models have been developed for a variety of aircraft types including manned aircraft, large unmanned aircraft systems (UAS), unconventional aircraft such as helicopters and balloons, and small hobbyist drones \cite{kochenderfer2010airspace,mueller2016simulation}. These models take the form of dynamic Bayesian networks that were trained from a large collection of data. The encounter model for manned aircraft, for example, was based on nine months of radar data covering much of the continental United States and containing almost 400,000 encounters \cite{kochenderfer2008correlated, kochenderfer2010airspace}. Recent work has focused on terminal airspace modeling, where Gaussian mixture models (GMMs) are used to represent distributions over trajectories \cite{barratt2018learning}. 

\begin{figure*}[htb]
    \centering
    \input{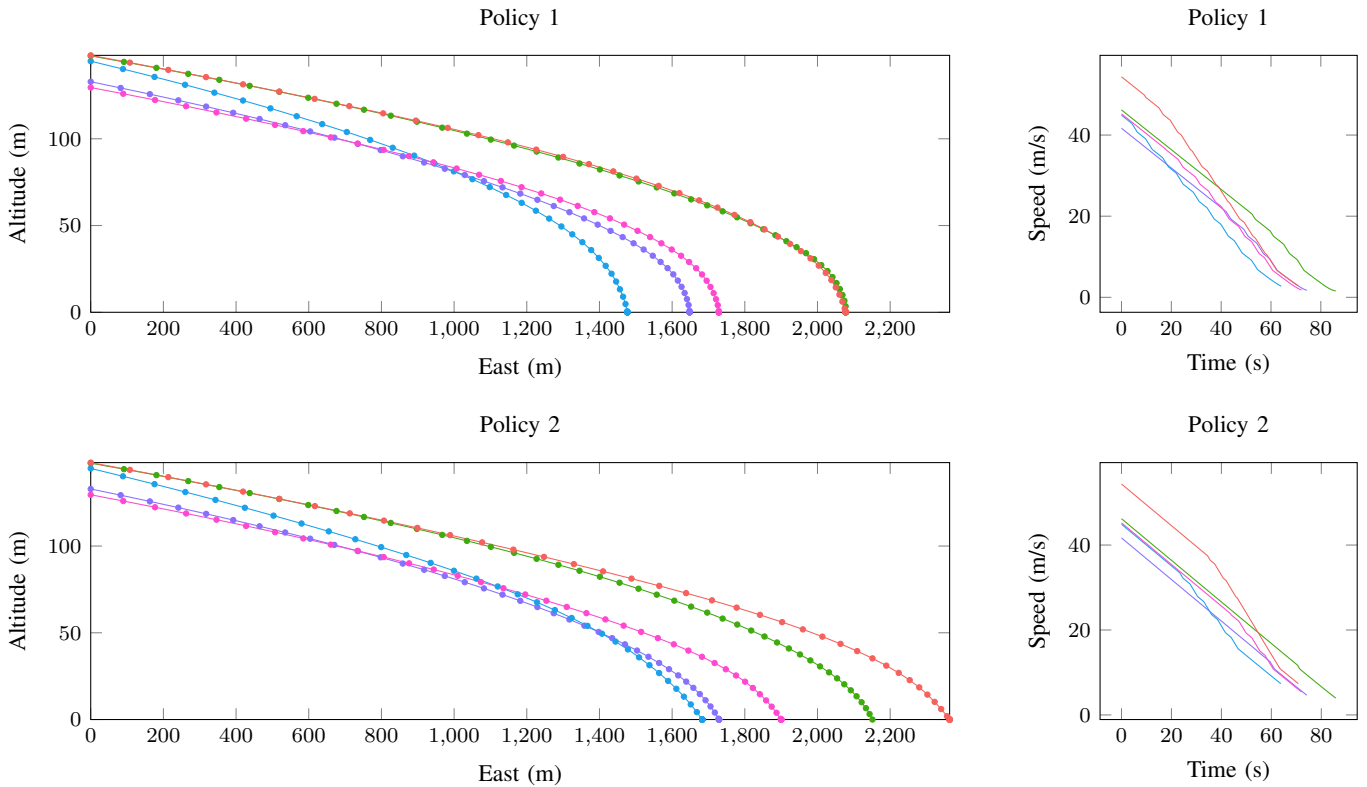}
    \caption{Sample query. Plots are generated by sampling initial states and performing rollouts of the query policies. The initial states are the same for each policy. The expert is asked to choose the policy that appears more realistic. \label{fig:samplequery}}
\end{figure*}

A significant challenge in building models for UAM is the lack of available data since such systems are not yet deployed. Without the ability to rely on data, a new approach to aircraft encounter modeling is needed that uses expert knowledge. This paper extends recent advances in preference-based learning to translate expert knowledge into a statistical encounter model that is expected to be representative of the future airspace. We develop an approach that learns an expert's preferences over potential encounter models through pairwise queries and demonstrate that this approach can be used to generate realistic trajectories in only a few minutes of an expert's time. \Cref{fig:samplequery} shows an example of a query used to elicit preferences over UAM landing trajectories.

This paper is organized as follows. \Cref{relwork} discusses related work. \Cref{approach} outlines our approach. \Cref{landing} describes an application of this approach to generate realistic UAM landing trajectories. \Cref{results} summarizes experimental results. \Cref{conclusion} concludes and recommends areas of future work.

\section{Related Work} \label{relwork}
Preference-based learning involves learning a policy or reward function directly from an expert's preferences. Its motivation is similar to that of inverse reinforcement learning (IRL) in that it allows experts to tune a policy without explicitly specifying the values of the reward function parameters \cite{ng2000algorithms}. A key difference between the two methods, however, is that preference-based learning does not require the user to provide demonstrations of optimal trajectories, which can be difficult in complicated, high-dimensional problem settings. For this reason, preference-based learning has been successfully applied to challenging problems in robotics \cite{akrour2011preference, wilson2012bayesian, christiano2017deep, dorsa2017active, biyik2018batch}. Preference-based learning has also been used in multiobjective design optimization \cite{alg4opt}. Developing a robust collision avoidance system requires optimizing over multiple metrics that have inherent trade-offs (e.g.\, maximizing safety while minimizing unnecessary alerts). During the development of ACAS X, preference-based learning was applied to elicit expert preferences over the balance between these competing objectives \cite{lepird2015bayesian}.

There are multiple considerations in selecting a preference-based learning framework. For instance, both non-Bayesian \cite{alg4opt, iyengar2001evaluating} and Bayesian \cite{wilson2012bayesian, lepird2015bayesian, dorsa2017active, biyik2018batch} approaches have been proposed in the literature. 
Preferences can be used to either learn a policy directly \cite{wilson2012bayesian, christiano2017deep} or indirectly by first learning a reward or utility function \cite{dorsa2017active, biyik2018batch, lepird2015bayesian, alg4opt}. Learning a reward or utility function typically involves a lower-dimensional parameter space and makes it easier to incorporate prior knowledge; however, it requires the additional step of translating the learned reward into a policy. 

Preference-based learning algorithms require an iterative process between consulting the expert and updating the model to select the next query. Optimal querying methods have been well-studied in the past. In the deterministic case, the Q-Eval algorithm selects design points that maximally reduce the volume of the set of feasible model parameters based on the current set of preferences \cite{iyengar2001evaluating}. A similar approach for the Bayesian case has been proposed, in which the algorithm seeks to maximize the minimum volume removed from the probability distribution over model parameters \cite{dorsa2017active}. Using this approach involves actively generating query trajectories by optimizing over a continuous action space. The efficiency of this approach was later improved upon by selecting queries in batches from a discrete sampling of trajectories \cite{biyik2018batch}. Other work compares two heuristic approaches that operate on a discrete set of trajectories \cite{wilson2012bayesian}. The first approach seeks to select the two trajectories that maximally disagree while still remaining likely, while the second approach selects queries based on the expected belief change. 

\section{Approach} \label{approach}
As outlined by \citeauthor{kochenderfer2010airspace}, encounter models may be correlated or uncorrelated in their structure \cite{kochenderfer2010airspace}. A correlated encounter model captures the effect of intervention from air traffic control when aircraft fly near one another, resulting in a correlation between the encounter trajectories. Uncorrelated models, on the other hand, model aircraft trajectories independently. A UAM vehicle may come across a number of possible intruders in the airspace such as hobbyist unmanned aircraft systems (UAS), commercial UAS, helicopters, and general aviation aircraft. Because of the lack of data representing typical aircraft behavior for UAM vehicles in close proximity with these other aircraft types, we choose an uncorrelated model in which the trajectories of intruder aircraft and UAM vehicles are generated independently before combining them to create an encounter. This approach allows us to use existing models for the other aircraft types. We are therefore focused on modeling realistic UAM vehicle trajectories, specifically at low altitudes where these vehicles are expected to be following landing and takeoff procedures.

Our model takes the form of a stochastic policy in a Markov decision process with preferences (MDPP) \cite{wirth2017survey}. In an MDPP, the reward function is not explicitly specified but rather learned from expert preferences. Once the reward function has been learned from experts, we are left with a traditional Markov decision process (MDP) that can be solved using dynamic programming to obtain a stochastic policy \cite{DMU}. Using this policy, trajectories can be easily sampled from the model as with prior encounter models. 

Applying preference-based learning in this problem setting requires multiple design considerations and adaptations of previously developed algorithms. For example, instead of trying to learn a policy that accomplishes a specific task, we are trying to find a policy that performs realistically in a wide range of scenarios. For this reason, the pairwise queries we present to the experts need to provide a comparison of the performance of \textit{policies} rather than \textit{trajectories}.
Queries are generated from policies obtained by solving the MDP for two different reward functions based on samples from our current estimate of the reward function parameters. 

Another key challenge that our method overcomes is the necessity of encoding prior knowledge into the queries. Encoding prior knowledge is especially important for our application since UAM vehicles at low altitudes (below 500 ft) will be flying with a specific intent (i.e.\, following takeoff or landing procedures). If we do not encode this prior knowledge, we may waste queries learning information we already know. For example, if we know that we are trying to model UAM landing trajectories, we should not present policies that do not result in a landing. We achieve this by allowing the expert to tune only some of the reward parameters with their preferences while holding others, such as a high reward for landing, fixed.

\subsection{Markov Decision Process with Preferences}
An MDP is a way of encoding a sequential decision making problem where an agent's action at each time step depends only on its current state \cite{DMU}. An MDP is defined by the tuple $(S, A, T, R, \gamma)$, where $S$ is the state space, $A$ is the action space, $T(s,a,s')$ is the probability of transitioning to state $s'$ given that we are in state $s$ and take action $a$, $R(s,a)$ is the reward for taking action $a$ in state $s$, and $\gamma$ is the discount factor. In this work, we assume that $R(s,a)$ is defined by a set of parameters (e.g.\, acceleration penalty and landing reward), some with fixed values and the others with values that are to be learned from the expert. We denote the vector containing parameters to be learned from the expert as $\mathbf{w}$ and call the resulting reward function $R_\mathbf{w}(s,a)$. We narrow our search space by restricting $\mathbf{w}$ such that $\| \mathbf{w} \|_1 = 1$ and requiring that the parameters in $\mathbf{w}$ have an implicit trade-off.

This reward function allows us to generate a policy that maps from states to actions. For any particular reward function, we can define a value $Q_\textbf{w}(s,a)$ associated with taking action $a$ from state $s$, with higher values indicating higher expected reward. With a discrete state and action space, the optimal value for each state and action, $Q_\textbf{w}^\ast(s,a)$, can be obtained using a form of dynamic programming called value iteration. The algorithm relies on iterative updates of $Q_\textbf{w}^\ast(s,a)$ using the Bellman equation \cite{bellman1954theory}:
\begin{equation}
    Q_\textbf{w}^\ast(s,a) = R_\mathbf{w}(s,a) + \sum_{s' \in S} T(s,a,s') \max_{a' \in A} Q_\textbf{w}^\ast(s',a')
\end{equation}
A deterministic policy $\pi(s)$ can be defined by simply choosing the action with the maximum value at state $s$:
\begin{equation}
    \pi(s) = \argmax_{a \in A} Q_\textbf{w}^\ast(s,a)
    \label{detpol}
\end{equation}

In order to evaluate the robustness of UAM systems, we need a probabilistic model of aircraft trajectories. Therefore, we use a stochastic policy in the form of a softmax policy with precision parameter $\lambda$ that controls the randomness of our actions. The probability we select action $a$ from state $s$ is given by
\begin{equation}
    p(a \mid s) = \frac{\exp[\lambda Q_\textbf{w}^\ast(s,a)]}{\sum_{a' \in A}\exp[\lambda Q_\textbf{w}^\ast(s,a')]}
    \label{eq:stochasticpol}
\end{equation}
As $\lambda \to \infty$, the policy approaches the one defined by \cref{detpol}. As $\lambda \to 0$, we approach a policy that chooses actions uniformly at random.

\subsection{Preference Model}
Because it is difficult for an expert to perfectly distinguish realistic trajectories, we adopt a Bayesian approach to preference modeling that allows errors. The mathematical framework and notation used in this work are based on the model of \citeauthor{dorsa2017active} \cite{dorsa2017active}. We keep a distribution over the possible values of $\mathbf{w}$ and perform Bayesian updates to it as we obtain responses to queries. We will call this distribution $p(\mathbf{w})$. The $n$th query consists of two sets of trajectories, $\tau^{(n)}_a$ and $\tau^{(n)}_b$. Let $I_n$ be the response to the $n$th query, where
\begin{equation}
    I_n = \begin{cases}
    +1, & \tau^{(n)}_a \succ \tau^{(n)}_b \\
    -1, & \tau^{(n)}_a \prec \tau^{(n)}_b
    \end{cases}
\end{equation}
with $\tau_a \succ \tau_b$ representing the expert's preference of $\tau_a$ to $\tau_b$. The Bayesian update can be written as follows:
\begin{equation}
    p(\mathbf{w} \mid I_n) \propto p(I_n \mid \mathbf{w}) p(\mathbf{w})
\end{equation}
where $p(\mathbf{w})$ encodes the current distribution over $\mathbf{w}$ that takes into account responses $I_{1:n-1}$. Before any preferences have been obtained, we assume a uniform prior over the search space of possible values.

In order to perform this update, we must specify a likelihood model for $p(I_n \mid \mathbf{w})$ keeping in mind that we expect occasional errors from the expert. As in \citeauthor{dorsa2017active} \cite{dorsa2017active}, we use a sigmoid likelihood function:
\begin{equation}
    p(I_n \mid \mathbf{w}) = \frac{1}{1 + \exp[-I_n(R_\mathbf{w}(\tau_a^{(n)}) - R_\mathbf{w}(\tau_b^{(n)}))]}
\end{equation}
We overload our notation for the reward function to allow $R_\mathbf{w}(\tau)$ to represent an evaluation of the reward of a particular trajectory set $\tau$ given the parameters $\mathbf{w}$. It is worth noting that the definition of $R_\mathbf{w}(\tau)$ can have a significant effect on algorithm performance. It is important that this function accurately encode the trade-off between the features that determine the reward function. One possible definition is the average reward over all trajectories in the set as shown in the example provided in this work, but feature tuning may be required and other definitions may provide better performance. If query policies are made stochastic, $R_\mathbf{w}(\tau)$ will need to be based on the likelihood of the trajectories in the set according to the policy defined by $\mathbf{w}$ and precision parameter $\lambda$. This approach is left for future work.
The posterior probability $p(\mathbf{w} \mid I_n)$ can be estimated using Markov Chain Monte Carlo (MCMC) methods. In particular, we use the adaptive Metropolis algorithm to efficiently generate samples at each iteration \cite{haario2001adaptive}.

\subsection{Querying Methods} \label{queries}
Bayesian methods for generating queries to learn reward functions have previously been proposed \cite{dorsa2017active,biyik2018batch}. One approach is to generate queries by solving an optimization problem over a continuous control input that maximizes the volume removed from $p(\mathbf{w})$ with the expert's response. However, while the optimization problem provides incentive to generate query trajectories that are maximally informative, it does not directly provide incentive to generate query trajectories that are realistic. This issue is especially apparent when there is significant prior knowledge of how the agent should perform. In the case of generating a realistic landing trajectory, we know that we want the aircraft to land, but it is difficult to encode this knowledge into the optimization problem. 

Instead, we generate queries by performing a specified number of rollouts of two policies obtained by solving the MDP with different vectors $\mathbf{w}$. These vectors are chosen from the MCMC samples used to estimate $p(\mathbf{w})$. The following querying methods provide heuristics for choosing informative samples of $\mathbf{w}$ to use in the queries.

\subsubsection{Multiobjective Optimization}
The first method relies on a multiobjective optimization problem that is similar to the query by disagreement method of \citeauthor{wilson2012bayesian} \cite{wilson2012bayesian}. Let $M$ be the number of MCMC samples generated from each Bayesian update, and let $\mathbf{w}_i$ and $\mathbf{w}_j$ with $i,j \in \{1,\ldots,M\}$ be the $i$th and $j$th samples, respectively. The optimization problem can be written as, 
\begin{equation}
    \underset{i,j \text{ s.t. } i \neq j}{\text{maximize }} p(\mathbf{w}_i)p(\mathbf{w}_j) + \mu \| \mathbf{w}_i - \mathbf{w}_j \|_2
\end{equation}
where $\mu \geq 0$ controls the balance between the objectives. This can be thought of as a heuristic balance between exploration and exploitation. The first term incentivizes selecting parameters that are likely based on the current estimate of the reward function, while the second ensures that the samples differ enough to allow for a comparison. We estimate $p(\mathbf{w}_i)$ using a Gaussian kernel density estimate based on the $M$ samples of $\mathbf{w}$.

\subsubsection{Probabilistic Q-Eval}
The second method adapts the Q-Eval algorithm \cite{iyengar2001evaluating} to a probabilistic setting. In the deterministic case where expert preferences are consistent, it is possible to compute a permissible region of potential values of $\mathbf{w}$. Q-Eval seeks to maximally reduce the size of this region with every query by choosing the samples that come closest to bisecting it. We extend this to the probabilistic case by approximating the permissible region using our samples. The steps for selecting a query are as follows:
\begin{enumerate}
    \item We use the sample mean to estimate the center of the permissible region: $\mathbf{c} = \frac{1}{M}\sum_{i=1}^M\mathbf{w}_i$.
    \item We compute the normal distance from the bisecting hyperplane between each pair of samples to $\mathbf{c}$.
    \item For each of the $k$ bisecting hyperplanes closest to $\mathbf{c}$, we estimate the volume ratio between both sides of the hyperplane based on the number of samples on each side.
    \item We create the next query by selecting the pair of samples corresponding to the hyperplane with a ratio of samples on either side closest to 1.
\end{enumerate}


\algnewcommand{\Initialize}[1]{%
  \State \textbf{Initialize:}
  \Statex \hspace*{\algorithmicindent}\parbox[t]{.8\linewidth}{\raggedright #1}
}

\subsection{Algorithm Summary}
Our method for learning the MDP reward parameters is summarized in \cref{alg:ri}. We begin with an empty set of preferences and generate our initial samples of $p(\mathbf{w})$ from a uniform distribution over the search space. On each iteration, we select two of our current samples from $p(\mathbf{w})$ using one of the query selection methods. The resulting reward functions, $R_{\mathbf{w}_1}(s,a)$ and $R_{\mathbf{w}_2}(s,a)$, are used to create policies for the next query by solving the MDP. After obtaining a random sampling of initial states, the next query is generated by performing rollouts of each policy from our sampled initial states. Finally, the expert's preference is obtained and used to update $p(\mathbf{w})$ through MCMC sampling. This process is repeated until we obtain a final estimate of the reward function.

\begin{algorithm}
\caption{Reward Iteration\label{alg:ri}}
\begin{algorithmic}[1]
\Function{reward\_iteration}{max\_iter, $M$}
    \State prefs $\gets \emptyset$
    \State $\mathbf{w}_\text{samples} \gets M$ samples from uniform prior
    \For{$i\gets 1$ to max\_iter}
       \State $\mathbf{w}_1, \mathbf{w}_2 \gets \text{select\_query\_reward}(\mathbf{w}_\text{samples}$)
       \State $\pi_1, \pi_2 \gets$ solve\_mdp($\mathbf{w}_1, \mathbf{w}_2$)
       \State init\_states $\gets$ samples from initial state distribution
       \State query $\gets$ generate\_query($\pi_1, \pi_2$, init\_states)
       \State prefs $\gets$ prefs $\cup$ obtain\_preference(query)
       \State $\mathbf{w}_\text{samples} \gets$ MCMC(prefs)
    \EndFor
    \State \textbf{return} estimate\_w($\mathbf{w}_\text{samples}$)
\EndFunction
\end{algorithmic}
\end{algorithm}

\section{UAM Landing Example} \label{landing}
The methods developed in the previous section were applied to generate realistic landing trajectories for UAM vehicles. The following subsections outline our design choices. An implementation can be found at https://github.com/sisl/UAMPreferences.

\subsection{State and Action Space}
Since the MDP must be solved twice during each query generation step, state variables and their discretizations were selected to be maximally expressive while still allowing the problem to remain tractable. \Cref{tab:statespace} summarizes the state variables.
\begin{table}[htb]
    \centering
    \caption{State Variables \label{tab:statespace}}
    \begin{tabular}{@{}llr@{}}
         \toprule
         \textbf{State Variable} &  \textbf{Description} & \textbf{Number of Values} \\
         \midrule
         $h$ & altitude above ground level & 50 \\
         $\dot{h}$ & vertical rate & 4 \\
         $\dot{x}$ & ground speed & 15 \\
         $a_{\text{prev}}$ & previous action & 16 \\
         \bottomrule
    \end{tabular}
\end{table}
Adding the previous action to our current state space allows us to penalize a change in acceleration (jerk) while still satisfying the Markov property.

At each timestep, the vehicle selects both a vertical and horizontal acceleration. We select 4 possible values for each type of acceleration (one of which is zero) resulting in 16 total joint actions. Actions were chosen so that small changes in the reward function will still result in different policies. The transition between states for a particular action is deterministic and can be calculated using kinematics.

\subsection{Reward Function}
Our reward function has five parameters that together promote a soft landing. Of the five parameters, two are fixed and three are left to be tuned by expert preferences. \Cref{tab:rewardparams} outlines the parameters and their descriptions. We denote the value for unknown parameters with a question mark.
\begin{table}[htb]
    \centering
    \caption{Reward Function Parameters \label{tab:rewardparams}}
    \begin{tabular}{@{}llr@{}}
         \toprule
         \textbf{Parameter} &  \textbf{Description} & \textbf{Value} \\
         \midrule
         $\ell$ & landing reward & $10000$ \\
         $b$ & penalty for flying backwards & $-0.1$ \\
         $\alpha$ & jerk penalty & ? \\
         $\beta$ & penalty for speed near the ground & ? \\
         $\gamma$ & acceleration penalty & ? \\
         \bottomrule
    \end{tabular}
\end{table}
The first parameter, $\ell$, allows us to ensure that all of the trajectories presented in the queries will result in a landing, so its value is kept at a large, fixed number. The second parameter, $b$, was added to penalize the vehicle for slowing down so much that it starts moving backwards, which was observed during initial tests. 

For this particular scenario, the vector $\mathbf{w}$, defined as $\mathbf{w} = [\alpha, \beta, \gamma]^\top$, contains our unknown parameters. Together these parameters define $R_\mathbf{w}(s, a)$. The parameters $\alpha$ and $\gamma$ penalize the jerk and acceleration respectively, while $\beta$ is meant to encourage a soft landing. The speed near the ground is wrapped into a feature, $\phi_\beta$, that gets multiplied by $\beta$ to determine this portion of the reward as follows:
\begin{equation}
    \phi_\beta = \begin{cases}
    -\min \left(\|  [ \dot{h},  \dot{x} ] \|_2 /h, 1000 \right), & h < h_{pen} \\
    0, & \text{otherwise}
    \end{cases}
\end{equation}
where $h$, $\dot{h}$, and $\dot{x}$ are all provided in the current state $s$ and $h_{pen}$ is a fixed parameter. We choose $h_{pen}$ to be 50 ft. The magnitude of the penalty for any particular speed increases as the vehicle gets closer to the ground.

As mentioned previously, we require that $\| \mathbf{w} \|_1 = 1$ in order to narrow our search space. In this particular application, we also know the polarity of each element of $\mathbf{w}$ using our prior knowledge that each parameter represents a penalty. Thus, we choose negative features of our state and action and restrict the elements of $\mathbf{w}$ to be strictly positive. It is important to note that the features of the state and action that are multiplied by our reward function parameters at each time step such as acceleration, jerk, and landing speed should be normalized so that they are similar in magnitude. This normalizations allows the reward function parameters to also have similar magnitudes to one another and increases the efficiency of our sampling.

We define our function $R_\mathbf{w}(\tau)$ to be the average reward over all trajectories. We use average reward in this scenario because some trajectories require more time to land than others, so we are comparing trajectories of different lengths in our queries. More reward function feature tuning was performed at this stage in order to normalize average reward and preserve the trade-off between model parameters.  

\section{Results} \label{results}
The algorithm's performance was evaluated by answering the preference queries according to a true hidden reward function defined by $\mathbf{w}_\text{true}$ that was selected for demonstration purposes. In order to assess the algorithm's progress, we rely on the cosine similarity metric used in prior work \cite{dorsa2017active, biyik2018batch}:
\begin{equation}
    \text{cosine similarity} = \mathbb{E}\left[\frac{\mathbf{w} \cdot \mathbf{w}_\text{true}}{\|\mathbf{w}\|_2 \|\mathbf{w}_\text{true}\|_2}\right]
\end{equation}
This metric provides a way to measure how close our estimate is to the true reward function at any given iteration. Higher values indicate better estimates. An exact match between the estimate and true value would result in a value of 1.

\subsection{Parameter Tuning}
\begin{figure}[htb]
    \centering
    \input{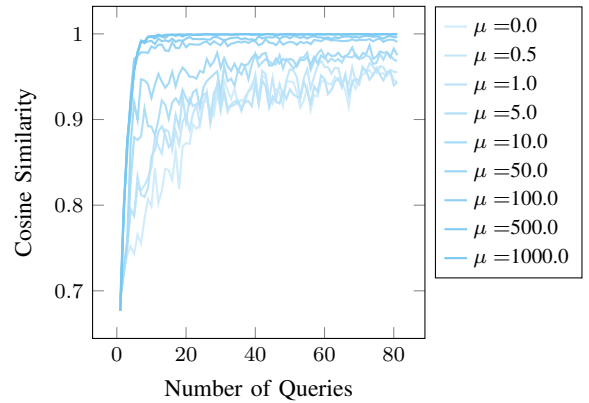}
    \caption{Performance of multiobjective optimization querying method for varying values of $\mu$. All curves are averaged over five trials. \label{fig:multiobj tuning}}
\end{figure}

The multiobjective optimization querying method is sensitive to the hyperparameter $\mu$. For this reason, we analyzed the algorithm performance over a range of possible values. \Cref{fig:multiobj tuning} shows the results averaged over five trials.
The algorithm performs poorly for low values of $\mu$. When $\mu$ is small, the selected samples are likely too close to one another to produce an informative query. A larger value for $\mu$ promotes a more efficient exploration of the parameter space and therefore faster convergence. Performance improves each time we increase $\mu$ but reaches a limit around $\mu=500$. All subsequent analyses were run with $\mu=500$. These results illustrate the importance of selecting query policies that maximally differ from one another.

\subsection{Overall performance}
\begin{figure}[htb]
    \centering
    \begin{tikzpicture}[]
\begin{axis}[height = {8cm}, legend pos = {south east}, ylabel = {Cosine Similarity}, xlabel = {Number of Queries}, width = {8cm}, legend style={font=\footnotesize}, ylabel near ticks]\addplot+ [mark = {none}, solid, thick, pastelBlue, mark=none]coordinates {
(1.0, 0.6767352840647222)
(2.0, 0.7899372245777825)
(3.0, 0.8736770410058604)
(4.0, 0.9196417023750427)
(5.0, 0.9572400521528582)
(6.0, 0.9788671893158837)
(7.0, 0.9920287046942627)
(8.0, 0.9892995088948308)
(9.0, 0.9962697092151969)
(10.0, 0.9965622958707282)
(11.0, 0.9945929087831435)
(12.0, 0.9976554610084145)
(13.0, 0.9968490200899327)
(14.0, 0.9954496652179061)
(15.0, 0.9985782460685184)
(16.0, 0.9988730035613262)
(17.0, 0.997756984348355)
(18.0, 0.9988852255867418)
(19.0, 0.9968451459291068)
(20.0, 0.9991183484947921)
(21.0, 0.9993865832103024)
(22.0, 0.9987138616208568)
(23.0, 0.9988143711091798)
(24.0, 0.9985562679735345)
(25.0, 0.9993825175375349)
(26.0, 0.9992419478122672)
(27.0, 0.9993036660059792)
(28.0, 0.9991989053959042)
(29.0, 0.9989547396165438)
(30.0, 0.9995097509105728)
(31.0, 0.9992927117859234)
(32.0, 0.9994486665905085)
(33.0, 0.9990784916380339)
(34.0, 0.9990513054432888)
(35.0, 0.9993671145439109)
(36.0, 0.9994399653180169)
(37.0, 0.9997106175583583)
(38.0, 0.9996131477054767)
(39.0, 0.9988169661643826)
(40.0, 0.9990388141589752)
(41.0, 0.9994233967446263)
(42.0, 0.9985817262202559)
(43.0, 0.9995473378460108)
(44.0, 0.9997711533259152)
(45.0, 0.9993418850652649)
(46.0, 0.9988808518785355)
(47.0, 0.9989863284496447)
(48.0, 0.999707649026034)
(49.0, 0.9996568118283793)
(50.0, 0.998815939134472)
(51.0, 0.9997982218865662)
(52.0, 0.9989733357463763)
(53.0, 0.9993363744716222)
(54.0, 0.9993521606677491)
(55.0, 0.9992546134195507)
(56.0, 0.9995883406813825)
(57.0, 0.9984059254885791)
(58.0, 0.9996368901490953)
(59.0, 0.9992469839424801)
(60.0, 0.9994484700537584)
(61.0, 0.9997049466361367)
(62.0, 0.9993622921499096)
(63.0, 0.9995271574653994)
(64.0, 0.9995087448983597)
(65.0, 0.9992548427712282)
(66.0, 0.9995236536641962)
(67.0, 0.9995858341094939)
(68.0, 0.9995063867022516)
(69.0, 0.999488723563432)
(70.0, 0.9996733485280777)
(71.0, 0.9997577776130843)
(72.0, 0.9993478107926886)
(73.0, 0.9993255015428048)
(74.0, 0.9989804893056371)
(75.0, 0.9994814591190624)
(76.0, 0.9996096277922628)
(77.0, 0.9997298241152365)
(78.0, 0.9997166478120748)
(79.0, 0.9997047377554076)
(80.0, 0.999598415674552)
(81.0, 0.9992812516371522)
};
\addlegendentry{Multiobjective}
\addplot+ [mark = {none}, solid, thick, pastelRed, mark=none]coordinates {
(1.0, 0.689764505030145)
(2.0, 0.8112040629258048)
(3.0, 0.839687764791768)
(4.0, 0.8810765026232025)
(5.0, 0.9249669957590253)
(6.0, 0.9242352718691738)
(7.0, 0.9390124011051)
(8.0, 0.9403540322782383)
(9.0, 0.9530589782859815)
(10.0, 0.9499760743486313)
(11.0, 0.9582958558842443)
(12.0, 0.9718446144832009)
(13.0, 0.9615198876054146)
(14.0, 0.9619205319860384)
(15.0, 0.9803918522639632)
(16.0, 0.9804386009120352)
(17.0, 0.9783838320707193)
(18.0, 0.9769649888640073)
(19.0, 0.9863425211793835)
(20.0, 0.9863391669822924)
(21.0, 0.9859047530031286)
(22.0, 0.9826381314678816)
(23.0, 0.9842451604419626)
(24.0, 0.9869738100277659)
(25.0, 0.9906526230557177)
(26.0, 0.9888375940374659)
(27.0, 0.9902792632919984)
(28.0, 0.9909392690752657)
(29.0, 0.9925545625722011)
(30.0, 0.9879338844323581)
(31.0, 0.9920962349518556)
(32.0, 0.9879154486643378)
(33.0, 0.9898968887506046)
(34.0, 0.9922181375390485)
(35.0, 0.9908566334854125)
(36.0, 0.9960704600143471)
(37.0, 0.9940899047108005)
(38.0, 0.9952516338099822)
(39.0, 0.996930519148641)
(40.0, 0.9972341926020668)
(41.0, 0.9929065365107601)
(42.0, 0.9969120482227909)
(43.0, 0.9978601851785864)
(44.0, 0.9947040949469255)
(45.0, 0.992600884958126)
(46.0, 0.9957469357216823)
(47.0, 0.9966724335635438)
(48.0, 0.9974088623177326)
(49.0, 0.9960141923569651)
(50.0, 0.9962047777337781)
(51.0, 0.996529989559893)
(52.0, 0.9970052529607191)
(53.0, 0.9961978829984913)
(54.0, 0.9975528382232526)
(55.0, 0.997101877319376)
(56.0, 0.996707300334767)
(57.0, 0.9978107564188653)
(58.0, 0.9978483002674613)
(59.0, 0.9984503453808544)
(60.0, 0.9980030131222071)
(61.0, 0.9984444402179926)
(62.0, 0.9979520846710688)
(63.0, 0.9978654735895034)
(64.0, 0.9975940395290526)
(65.0, 0.9972222762551898)
(66.0, 0.9988352788448847)
(67.0, 0.996156321826508)
(68.0, 0.9979228064610088)
(69.0, 0.9989175584920131)
(70.0, 0.9974645854039517)
(71.0, 0.9987391447190787)
(72.0, 0.9984622959010767)
(73.0, 0.9976623270183295)
(74.0, 0.9976760336328063)
(75.0, 0.9983035746460736)
(76.0, 0.9983521053026093)
(77.0, 0.9963322740136196)
(78.0, 0.9976052160586315)
(79.0, 0.9985539259756523)
(80.0, 0.9955480553590326)
(81.0, 0.9957336069390526)
};
\addlegendentry{Probabilistic Q-Eval}
\end{axis}

\end{tikzpicture}
    \caption{Comparison of convergence for each querying method. \label{fig:multvqeval}}
\end{figure}
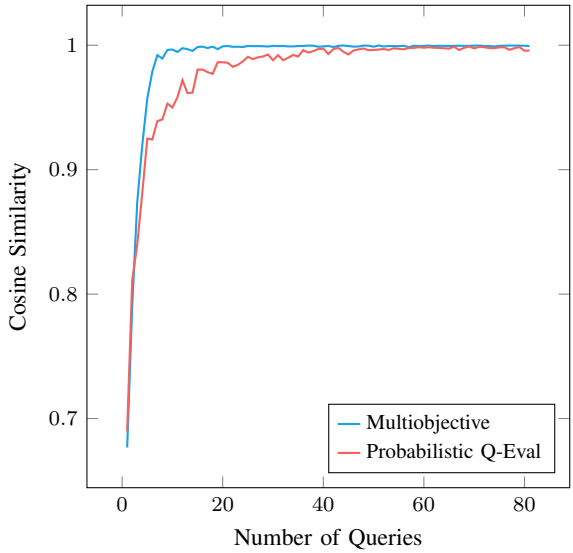

\Cref{fig:multvqeval} shows the algorithm performance using both the multiobjective and probabilistic Q-Eval querying methods. 
Both methods approach the true reward function, converging after about 40 queries. The properly tuned multiobjective optimization method approaches the true reward faster than the probabilistic Q-Eval method. Queries take under 15 seconds to generate on an single Intel Core i7 processor operating at 4.20 GHz. Accounting for the time it takes to respond, 40 queries translates to roughly 15 minutes of the expert's time. This process would take less time with parallelization.

\begin{figure}[htb]
    \centering
    \input{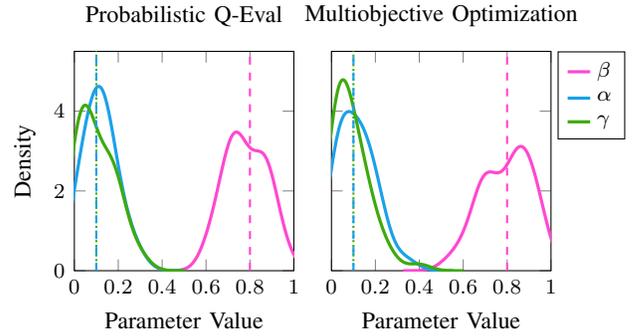}
    \caption{Univariate Gaussian kernel density estimate of the distribution of each parameter after 80 queries. Vertical lines indicate the true values.\label{fig:1Dkde}}
\end{figure}

In addition to monitoring the changes in the parameter estimates as preferences are obtained, we also consider the progression of the distribution $p(\mathbf{w})$. \Cref{fig:1Dkde} shows the univariate Gaussian kernel density estimate of the final distribution for each model parameter. As expected, the final distributions are centered near the true values. 

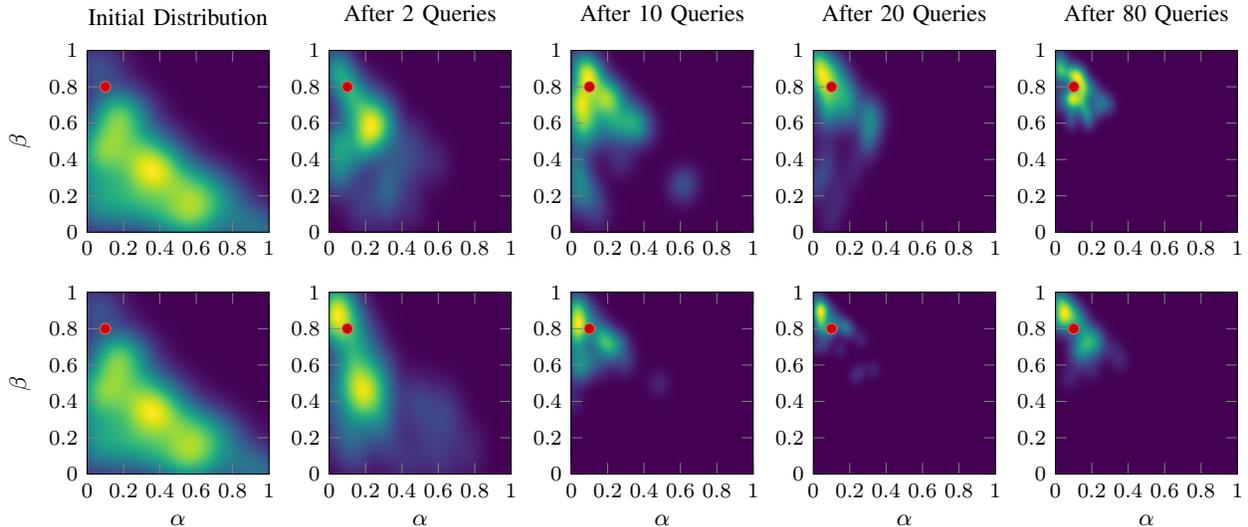
\begin{figure*}[htb]
    \centering
    \begin{tikzpicture}[]
\begin{groupplot}[xlabel=$\alpha$, ylabel=$\beta$, group style={xlabels at=edge bottom, ylabels at =edge left,
    horizontal sep=0.8cm, vertical sep=0.8cm, group size=5 by 2}]
\nextgroupplot [height = {4cm}, title = {Initial Distribution}, width = {4cm}, enlargelimits = false, axis on top]\addplot [point meta min=-2.478524764320075, point meta max=-2.8625646970676874e-13] graphics [xmin=0, xmax=1, ymin=0, ymax=1] {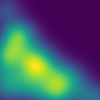};
\addplot+[draw=none, mark = {*}, pastelRed] coordinates {
(0.1, 0.8)
};
\nextgroupplot [height = {4cm}, title = {After 2 Queries}, width = {4cm}, enlargelimits = false, axis on top]\addplot [point meta min=-6.08880093920895, point meta max=-0.0] graphics [xmin=0, xmax=1, ymin=0, ymax=1] {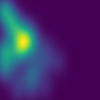};
\addplot+[draw=none, mark = {*}, pastelRed] coordinates {
(0.1, 0.8)
};
\nextgroupplot [height = {4cm}, title = {After 10 Queries}, width = {4cm}, enlargelimits = false, axis on top]\addplot [point meta min=-6.618445655764427, point meta max=-0.0] graphics [xmin=0, xmax=1, ymin=0, ymax=1] {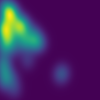};
\addplot+[draw=none, mark = {*}, pastelRed] coordinates {
(0.1, 0.8)
};
\nextgroupplot [height = {4cm}, title = {After 20 Queries}, width = {4cm}, enlargelimits = false, axis on top]\addplot [point meta min=-11.016279819937983, point meta max=-0.0] graphics [xmin=0, xmax=1, ymin=0, ymax=1] {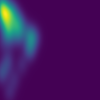};
\addplot+[draw=none, mark = {*}, pastelRed] coordinates {
(0.1, 0.8)
};
\nextgroupplot [height = {4cm}, title = {After 80 Queries}, width = {4cm}, enlargelimits = false, axis on top]\addplot [point meta min=-23.205702061615142, point meta max=-0.0] graphics [xmin=0, xmax=1, ymin=0, ymax=1] {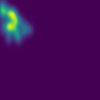};
\addplot+[draw=none, mark = {*}, pastelRed] coordinates {
(0.1, 0.8)
};
\nextgroupplot [height = {4cm}, width = {4cm}, enlargelimits = false, axis on top]\addplot [point meta min=-2.478524764320075, point meta max=-2.8625646970676874e-13] graphics [xmin=0, xmax=1, ymin=0, ymax=1] {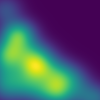};
\addplot+[draw=none, mark = {*}, pastelRed] coordinates {
(0.1, 0.8)
};
\nextgroupplot [height = {4cm}, width = {4cm}, enlargelimits = false, axis on top]\addplot [point meta min=-5.399227304762588, point meta max=-0.0] graphics [xmin=0, xmax=1, ymin=0, ymax=1] {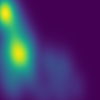};
\addplot+[draw=none, mark = {*}, pastelRed] coordinates {
(0.1, 0.8)
};
\nextgroupplot [height = {4cm}, width = {4cm}, enlargelimits = false, axis on top]\addplot [point meta min=-8.432438100136102, point meta max=-0.0] graphics [xmin=0, xmax=1, ymin=0, ymax=1] {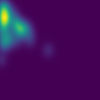};
\addplot+[draw=none, mark = {*}, pastelRed] coordinates {
(0.1, 0.8)
};
\nextgroupplot [height = {4cm}, width = {4cm}, enlargelimits = false, axis on top]\addplot [point meta min=-9.422992823648606, point meta max=-0.0] graphics [xmin=0, xmax=1, ymin=0, ymax=1] {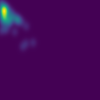};
\addplot+[draw=none, mark = {*}, pastelRed] coordinates {
(0.1, 0.8)
};
\nextgroupplot [height = {4cm}, width = {4cm}, enlargelimits = false, axis on top]\addplot [point meta min=-18.50335518491686, point meta max=-0.0] graphics [xmin=0, xmax=1, ymin=0, ymax=1] {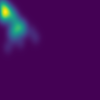};
\addplot+[draw=none, mark = {*}, pastelRed] coordinates {
(0.1, 0.8)
};
\end{groupplot}

\end{tikzpicture}
    \caption{Progression of bivariate Gaussian kernel density estimate of $p(\mathbf{w})$ as preferences are obtained. The red dot indicates the location of the target parameter values. The top row shows the progression for probabilistic Q-Eval, while the bottom row shows the progression for the multiobjective optimization method. \label{fig:2Dkde}}
\end{figure*}

\Cref{fig:2Dkde} shows the evolution of the bivariate Gaussian kernel density estimate for $\alpha$ and $\beta$. Due to our restrictions on the values in the vector $\mathbf{w}$, the reward function is completely defined by a selection of $\alpha$ and $\beta$. Furthermore, these parameters must lie in a two-dimensional simplex. We start with a uniform distribution over this simplex. As we obtain preferences, the distribution shrinks and centers itself near the true values.

\subsection{Effect of Human Error}
The results presented in the previous section are optimistic estimates since they do not take into account human error in answering the queries. For this reason, we conducted further experiments to determine the effect of human error on convergence. We again answer queries according to a predefined reward function, but we select the policy with a lower reward according to $\mathbf{w}_{\text{true}}$ with probability $\epsilon$. This probability can be thought of as the rate at which the expert responds to a query in a manner that is inconsistent with their true internal reward function. This model of human error is likely to be conservative since we are randomly choosing the queries that we answer incorrectly. In reality, an expert is more likely to provide an inconsistent answer to queries where both policies were similar than to queries that show significantly different policies. 

\begin{figure}[htb]
    \centering
    \input{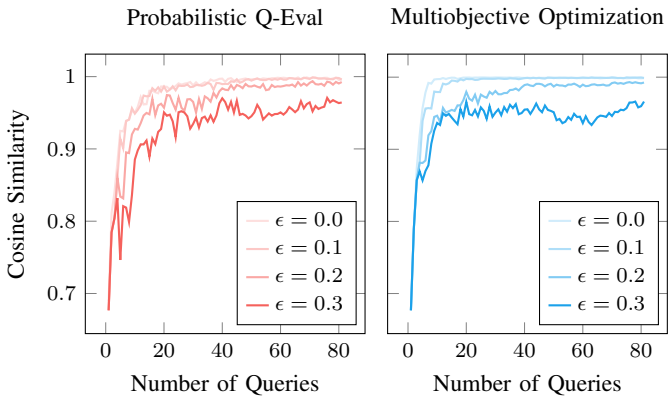}
    \caption{Error analysis for both querying methods. All curves are averaged over five trials. \label{fig:erroranalysis}}
\end{figure}

\Cref{fig:erroranalysis} shows the results of this analysis for both querying methods. 
Both methods show similar trends in performance as the error rate increases. Answering incorrectly at a rate of $\epsilon=0.1$ results in a minimal degradation in algorithm performance. When the error rate is increased to $\epsilon=0.2$, we begin to observe a noticeable effect in the number of queries required to converge; however, the algorithm still converges to an estimate close to the true value after approximately 40 queries. At an error rate of $\epsilon=0.3$, the algorithm no longer approaches the true reward.

\subsection{Stochastic Model}
\begin{figure}[htb]
    \centering
    \input{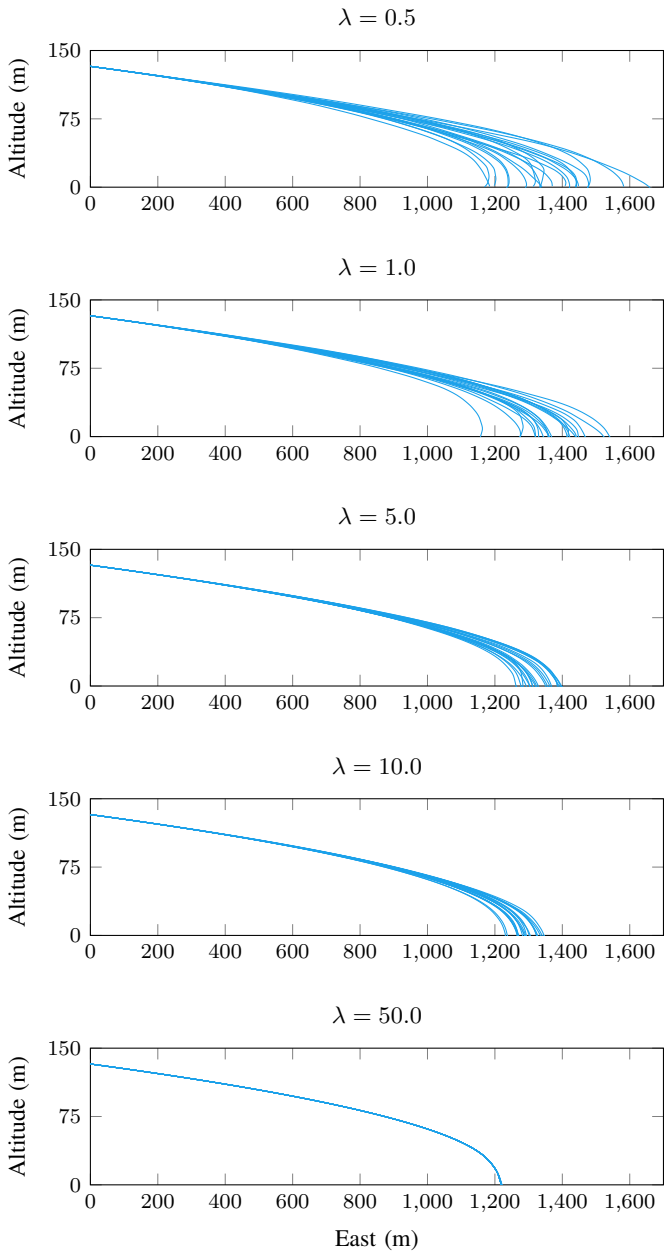}
    \caption{Effect of the precision parameter on the distribution of trajectories. Each plot contains 20 trajectories generated from the same initial state. \label{fig:stochasticexample}}
\end{figure}

In order to obtain a probabilistic model of aircraft trajectories, we need a stochastic policy. We use our final estimate of $\mathbf{w}$ to solve for $Q_\textbf{w}^\ast(s,a)$ for each state and action pair. We then perform trajectory rollouts using the policy defined in \cref{eq:stochasticpol} with a given precision parameter $\lambda$. \Cref{fig:stochasticexample} shows the effect of varying the precision parameter on the distribution of trajectories. The trajectories are generated from the same initial state using the optimal policy corresponding to $\mathbf{w}_\text{true}$.
Using this technique, the model can be sampled as needed to obtain encounter trajectories for simulation.

\section{Conclusion}\label{conclusion}
In this work, we have shown that the preferences of a domain expert can be used to generate realistic trajectories for UAM without depending on a large collection of data. Our method addresses two major challenges: allowing for a comparison of policies rather than trajectories and generating queries that incorporate prior knowledge. We overcome these challenges by learning only a subset of the reward parameters in an MDP from the expert. Our results show that we can achieve an accurate estimate of the model parameters after presenting a sensible number of queries to a domain expert. We also show that a Bayesian approach to preference modeling is robust to human error for error rates under 20\%. More broadly, this approach introduces a paradigm shift in the applications of preference-based learning that can be extended to other problem settings in which a realistic model must be created without a substantial dataset.

This work opens multiple paths for future work. For example, aligned with limitations noted in past work \cite{dorsa2017active,biyik2018batch}, determining effective features remains a difficult process. Features must accurately encode the trade-offs that an expert is internally optimizing over when they respond to a query. Future work will explore ways to select these features more efficiently. Furthermore, the model will better reflect the expert's true preferences if the stochastic nature of the resulting policies is illustrated in the queries rather than applied after training. Future studies will focus on ways to handle queries that show stochastic policies.

\section{Acknowledgements}
The authors would like to acknowledge Dorsa Sadigh, Erdem B{\i}y{\i}k, and Ransalu Senanayake for their helpful discussions throughout the progression of this work. This work was supported by $\text{A}^3$ by Airbus. Any opinions, findings, and conclusions expressed in this paper are those of the authors and do not necessarily reflect the views of $\text{A}^3$ by Airbus.

\renewcommand*{\bibfont}{\small}
\printbibliography

\end{document}